# Findings of MEGA: Maths Explanation with LLMs using the Socratic Method for Active Learning

Tosin Adewumi*, Foteini Simistira Liwicki, Marcus Liwicki, Viktor Gardelli⁶, Lama Alkhaled, Hamam Mokayed

Machine Learning Group, EISLAB, ⁶Education and Languages,

Luleå University of Technology, Sweden.

Abstract: This paper presents an intervention study on the effects of the combined methods of (1) the Socratic method, (2) chain-of-thought (CoT) reasoning, (3) simplified gamification and (4) formative feedback on university students' Maths learning driven by Large Language Models (LLMs). We call our approach Mathematics Explanations through Games by AI LLMs (MEGA). Some students struggle with Maths and as a result avoid Math-related discipline or subjects despite the importance of Maths across many fields, including signal processing. Oftentimes, students' Maths difficulties stem from suboptimal pedagogy. We compared the MEGA method to the traditional step-by-step (CoT) method to ascertain which is better by using a within-group design after randomly assigning questions for the participants, who are university students. Samples ($n = 60$) were randomly drawn from each of the two test sets of the Grade School Math 8K (GSM8K) and Mathematics Aptitude Test of Heuristics (MATH) datasets, based on the error margin of 11%, the confidence level of 90%, and a manageable number of samples for the student evaluators. These samples were used to evaluate two capable LLMs at length (Generative Pretrained Transformer 4o (GPT4o) and Claude 3.5 Sonnet) out of the initial six that were tested for capability. The results showed that students agree in more instances that the MEGA method is experienced as better for learning for both datasets. It is even much better than the CoT (47.5% compared to 26.67%) in the more difficult MATH dataset, indicating that MEGA is better at explaining difficult Maths problems. We also calculated the accuracies of the two LLMs and showed that model accuracies differ for the methods. MEGA appears to expose the hallucination challenge that still exists with these LLMs better than CoT. We provide public access to the MEGA app, the preset instructions[1] that we created and the annotations by the students for transparency.[2]

## I. INTRODUCTION: HISTORY, MOTIVATION, AND SIGNIFICANCE OF THE TOPIC

### A. History

Historically, many students struggle to understand Mathematics (Maths) and it has been argued that not all students will learn successfully when the same teaching method is applied [1]. A relatively recent survey by the Organization for Economic Cooperation and Development (OECD), for example, showed that only 13% of 15-year-old Swedish students scored above level 4 (on a scale from 2 to 6).[3] The OECD average was 11% while the percentage was slightly higher in some Asian countries (e.g. 21% in Korea). The scale is as defined by the Programme for International Student Assessment (PISA). Level 2 indicates students can model the way a simple situation may be represented mathematically (e.g. comparing the distances of alternative routes) while level 6 indicates they can model complex situations mathematically and evaluate their suitable problem-solving strategies.

Maths is an important part of many fields, such as astronomy, signal processing, data science, and generative artificial intelligence (GenAI), among others [2]–[4]. In ancient times in many cultures, Maths served as an important tool for preparing people for administration, such as in the training of scribes in Mesopotamia and Egypt [2]. Today, this is still true but with even broader purposes, such as explaining the physical world. Signal processing, which usually follows the system-centric approach, is a discipline that serves this broader purpose and is usually taught by using explicit Maths analysis to understand the physical world [3]. The system-centric approach involves the analysis and development of mathematically tractable systems for processing signals to ascertain the effects of external factors. Indeed, the two sub-fields of signal processing (analog and digital) are very mathematical, though analog signal processing usually requires more calculus [5].

---

[1] www.megamath.se, https://drive.google.com/drive/folders/1hal40Kaf4GvfLflPv6dKQ52bXljyFbQY?usp=sharing

[2] github.com/LTU-Machine-Learning/megamath

[3] oecd.org/content/dam/oecd/en/about/programmes/edu/pisa/publications/national-reports/pisa-2018/featured-country-specific-overviews/PISA2018_CN_SWE.pdf

*B. Motivation*

The introduction of technology (and GenAI) in education [6], generally, and the learning of Maths [7], in particular, present opportunities of new methods since traditional methods for explaining some concepts are inadequate for all students [1]. A Large Language Model (LLM) is a GenAI that is created using Machine Learning (ML) methods and is able to perform reasoning and dialog tasks to some degree of accuracy depending on the training and parameter size. More technically, an LLM is a probabilistic (neural network) model with a large number of parameters that is trained on large amounts of data to predict the next token (or word) from a probability distribution over a set of possible tokens [8]. The aim of this study is to investigate the capabilities of state-of-the-art (SotA) LLMs for Maths education in a university setting. Our **motivation** in this work is to answer the following **research question**: "*How do the learning experience of the quality of explanations to understand Maths differ between MEGA and the traditional CoT?*" The Socratic method, which is attributed to the Greek philosopher Socrates [9], is not new but the application of it in a simplified gamification context for Maths explanation, as done in this work, is new. Essentially, Socratic teaching is a form of dialogical education [10] with follow-up questions while the simplified gamification is the interactive engagement offered to users through multiple choice answers at each sub-question, where only one answer is correct. Two answer choices are implemented in this work.

*C. Approach and contributions*

To address our research question, we implemented Mathematics Explanations through Games by AI LLMs (MEGA), which combines (1) the Socratic method, (2) the traditional chain-of-thought (CoT), (3) simplified gamification, and (4) formative feedback. The method follows human-in-the-loop approach, where the LLMs do the work "with the students" in human evaluations that we carried out. We compared our approach with the CoT, which is the traditional step-by-step explanation, in pairwise comparison. A successful implementation of our approach (MEGA) has the potential impact of adapting the learning procedure for students with special needs, thereby providing a more inclusive, personalized and equitable education [11]. Furthermore, one of the main issues when using LLMs is that they suffer from "*hallucinations*" [12]. In some instances MEGA appears to offer some mitigation to this problem by the division of a Maths problem into simple and manageable sub-questions, thereby allowing the LLM to "think" or process the main question better. This does not offer a 100% solution to the hallucination problem, however, as will be noticed in this work. In the context of artificial intelligence (AI), the term hallucination is used when GenAI predicts false information confidently as fact.

Our approach involved conducting initial tests on six SotA LLMs to ascertain if they are capable of executing the MEGA method, which was designed as a system prompt. The six LLMs are Generative Pretrained Transformer 4o (GPT4o), Claude 3.5 Sonnet, Large Language Model Meta AI-3.1-70B-Instruct-Turbo (LLaMA-3.1-70B-Instruct-Turbo), Mistral Large, Cohere command-r-plus-08-2024, and Gemini 1.5 Flash. Only 3 (GPT4o, Claude 3.5 Sonnet, and LLaMA-3.1-70B-Instruct-Turbo) were capable of successfully following the system prompt and the first two were evaluated at length in this study after deploying them in our online app. From the test sets of two benchmark datasets (Grade School Math 8K (GSM8K) [13] and Mathematics Aptitude Test of Heuristics (MATH) [14]), we randomly sampled 60 Maths problems each, based on the error margin of 11%, the confidence level of 90%, and to have a manageable number of samples for student evaluators. The student evaluators are university students. The creators of the MATH dataset had also tested it with university students. In additional analysis, we calculated the accuracies of the two LLMs in three different settings: MEGA accuracies, Socratic (granular) accuracies, and CoT accuracies. The results showed that the MEGA method gives better learning experiences in the case of both datasets, according to the students. It is even much better than the CoT (47.5% compared to 26.67%) in the more difficult MATH dataset, indicating that MEGA is better at explaining difficult Maths problems. Furthermore, we (1) conducted content analyses of all the feedback from the evaluators, (2) showed that model accuracies differ among the methods, and (3) MEGA appears to expose the existing challenge of brittleness in the form of hallucination with these LLMs better than CoT. The significance of this work is evidenced from the **contributions** of our study. These include:

1) The introduction of the novel method Mathematics Explanations through Games by AI LLMs (MEGA), which, to the best of our knowledge, will be applied for the first time to the subject of Mathematics.
2) The public availability of our deployed application (MEGA) with multiple LLMs in the backend.
3) The provision of valuable insights on the capabilities of SotA LLMs and their limitations in implementing the MEGA method.

Hence, the novelty of MEGA is in combining the Socratic method and CoT with simplified gamification and formative feedback for Maths explanations, engineering the prompt (which is not native to ChatGPT or any other model) and creating the cloud app (which hosts several models, with at least 2 models with the capacity to execute the system prompt).

*D. Outline of the paper*

The rest of this paper is structured as follows. Section II identifies related studies in the literature using a tutorial outlook, including highlighting the ethical aspects of LLM usage in education. Section III discusses the details of our methodology. Section IV describes the results and various analyses. Section V concludes the paper with final remarks and possible directions for future studies.

## II. TUTORIAL-STYLE LITERATURE REVIEW

In the ancient Greco-Roman era, Maths education had a philosophical outlook [2]. By the 12th century, it had taken on more practical and commercial outlook with the establishment of universities and was taught as a secondary subject within the faculty of arts or medicine. In the 20th century, there were rapid advances in science and technology, fueled by Maths, among other factors. The opportunities for teaching Maths also grew during this period. Technology brought about the mass production of books, computers and the Internet [2]. This made it possible to have Maths evolve from being a subject for a few to a subject for all. In the following sub-sections, our **objective** is for the reader to be aware of the pedagogy of Maths and some of its content from a signal processing perspective, the Socratic method, and the SotA in Maths pedagogy tools.

*A. Mathematics and its pedagogy*

Three methods for teaching Maths are identified in the literature: (1) the traditional, (2) problem-solving, and (3) discovery learning [15]. The traditional method teaches by telling, thereby establishing the teacher (or instrument of delivery) as the focus. Typical lectures are examples of this method. It is entrenched in the **behaviorism** learning theory, which emphasizes changing behavior resulting from the learners' experience of associations of stimulus-response. The CoT method is situated in this method. The second method, problem-solving method, improves the reasoning skills of students by their active engagement in the solution process of the problem. It is entrenched in the **cognitivisim** learning theory. Four stages are identified in the problem-solving process by [15]: (1) having an understanding of the problem, (2) devising a plan on how to address the problem, (3) executing the plan with the option to revise it, if necessary, and (4) evaluating the results for correctness. Finally, the third method, the discovery method, makes students reflect on their mental and physical actions, thereby allowing them to gain new knowledge. It is entrenched in the **constructivist** theory, which emphasizes learner-centered learning. It ensures students think critically with significant guidance of the teacher by a social process that involves dialogue with themselves (inner monologue) as well as others or the teacher. The MEGA method is situated in this method. Overall, the three methods will have different success rates in the aim of meeting the needs of diverse students.

*B. Systematic synthesis of prior work from a signal processing perspective*

A signal, which is central to signal processing, is a function conveying information about an event or system. Signal processing focuses on analyzing, modifying and synthesizing signals. These signals can be digital or analog. Physical models of a system, often simplified, describe how data that may have been collected relates to useful information that may be needed. A common core of Maths ideas is useful for applications of signals from systems. For example, the Fourier transform and its variants (in the branch of transform theory) play important roles in many areas of signal processing. The Fourier transform, which expresses a signal as a combination of sine and cosine waves, is based on the Fourier theorem, which states that any periodic waveform can be described as the sum of a number of sinusoidal variations. According to [5], one can transform any arbitrary set of data into periodic components, even if it does not appear periodic. The generalization of the Fourier transform for discrete signals is called the z-transform. We also provide an example Laplace transform by MEGA in the appendix.

Signals may be sampled and when this is done at regular intervals it is called uniform sampling. The sampling theorem fundamentally states that a band-limited signal is uniquely specified by its sufficiently close, equally spaced samples, with the implication that it prevents the loss of any information despite allowing the replacement of a continuous band-limited signal by a discrete sequence of the samples. It establishes the theoretical equivalence of analogue and digital signals. The chart of discrete data points that are obtained from sampling a signal may be erratic and require smoothing (e.g. local averaging) to better assess the trend of the signal. The way continuous and discrete signals can be modeled and analyzed through Maths has evolved.

Expectedly, the content of what is taught in Maths in modern times is very different and more advanced than premodern times. Premodern times mainly involved the rudiments of arithmetic and geometry [2]. Today, there are many branches that are covered in Maths. These include **(1) algebra, (2) number theory, (3) trigonometry, (4) geometry, (5) calculus, (6) logic, (7) probability, (8) statistics, (9) information theory, and (10) numerical analysis**, among others. We hereby briefly describe these branches.

1) Algebra is a systematic notation of quantitative relationships involving unknown variables. For example, $y = 3x + 2$. A function is usually used to name this relationship involving the independent and dependent variables. It is useful in manipulating signals as vectors and matrices.

2) Number theory is a core foundation of Maths and involves the study of numbers, integers, and more. The set of natural numbers is the set that is used for counting, i.e. **N** = {0, 1, 2, …}, while the set of real numbers, **R**, is the set with numbers having decimal expression.
3) Trigonometry involves the relations of sides and angles of triangles. There are 3 basic trigonometric functions and 3 additional derived functions. For a given triangle with the opposite side, $O$, to the angle $\theta$, $A$, adjacent to the angle and $H$, the hypotenuse (longest side), the basic functions are $\sin \theta = \frac{O}{H}$, $\cos \theta = \frac{A}{H}$, and $\tan \theta = \frac{O}{A}$. The derived functions are cosecant $\theta = \frac{1}{\sin\theta}$, secant $\theta = \frac{1}{\cos\theta}$, and cotangent $\theta = \frac{1}{\tan\theta}$.
4) Geometry involves the study of shapes, angles, and spatial relationships. While signal processing has a tangential relation to geometry, it has a fundamental pillar of performing analysis and synthesis of functions as it uses sinusoids by working on vector spaces of signals. Most of the time, the basis are the orthogonal family of functions of the complex exponential. Sinusoids are used frequently in signal processing by Euler's identity: $e^{i\pi} + 1 = 0$, where $e$ is Euler's number, $i$ is the imaginary unit and $\pi$ is pi, the ratio of the circumference of a circle to its diameter.
5) Calculus involves the rate of change (differential calculus) and accumulation (integral calculus). The first derivative of a signal is the rate of change of the dependent variable, $y$, with $x$, the independent variable, which is the slope at each point of the tangent to the signal. Higher derivative orders can also be computed. It is useful in the analysis of continuous-time signals. Differentiation degrades signal-to-noise ratio, as is commonly observed, except it includes carefully optimized smoothing. On the other hand, integration fundamentally finds application as a method for calculating the dot product between two functions in a vector space of continuous functions. (or the average value of a signal).
6) Logic is the science of reasoning, evaluation of arguments and the conclusions to the arguments.
7) Probability involves the likelihood of an event happening. It is useful for modeling signals that have randomness.
8) Statistics involves separating and analyzing data to find patterns. Concepts useful to signal processing include hypothesis testing and statistical inference, among others.
9) Information theory involves quantifying the information in signals, which is useful for data compression and communication systems.
10) Numerical analysis provides approximation tools for solutions to Maths problems, particularly when complex computations are involved.

*C. The Socratic method and some fundamental theories*

Personalized education is more attainable now than ever before because of AI [16]. Combining the Socratic method in AI systems is another way to achieve this personalization, as carried out by [17]–[19]. However, their works did not involve Maths and **gamification** components, for the purpose of stimulating **active learning** or participation. Socrates is known for his dialogues with sub-questions to provoke critical thinking but not necessarily to arrive at an answer to an interlocutor's original main question [9]. His approach is firmly entrenched in active participation or learning - a method that has been proven to improve the learning of students because it is **student-centered** while also collaborative and dialogical [20]. AI-led explanations based on active learning contrasts with the inner monologue approach, where the "thinking" process of a model is hidden or kept internal from the student.

Active learning refers to instructional methods that actively engage students in the learning process through activities that require critical thinking, problem-solving, analysis, and reflection [21]. Unlike traditional lecture-based approaches, where students passively receive information, active learning emphasizes participation, collaboration, and interaction with the material, peers, and instructors. Active learning is mainly based on the concept that knowledge is not passively received but actively constructed by the learner. This strategy is based on constructivist principles, which let the learners engage deeply with content, work collaboratively, and apply their knowledge in meaningful contexts [22]. Furthermore, gamification in education aims to implement game design elements, mechanics, and principles in non-game educational settings to enhance student engagement, motivation, and learning outcomes [23]. Gamification in education is designed to align with the core psychological needs—autonomy, competence, and relatedness defined by the **self-determination theory (SDT)** [24]. The fulfillment of these core psychological needs will help to enhance intrinsic motivation and make the learning process more engaging and rewarding.

Student-centered learning is an educational approach that prioritizes the needs, interests, abilities, and learning preferences of students over the traditional teacher-centered model. In this paradigm, students take an active role in their learning process, engaging in decision-making about what, how, and when they learn. The teacher's role shifts from being the primary source of knowledge to a facilitator or guide who supports students in achieving their goals [25]. The theoretical foundations of student-centered learning are firmly built upon constructivist theory [22], SDT [24], and **experiential learning theory** [26]. Experiential learning theory posits that learning is most effective when it involves direct experience, critical reflection, and application of knowledge to real-world contexts. Together, these theories provide a robust foundation for designing educational environments where students are at the center, actively shaping their learning journey. With the help of natural language processing (NLP) and LLMs, educators can design student-centric environments where students actively engage with content, make choices about their learning pathways, and receive individualized support tailored to their unique needs and preferences [27].

It was shown by [28] that individualized, student-centered learning (or one-to-one tutoring) caused students to outperform a control group by two standard deviations. This is known as the **two sigma (or two standard deviations) problem**. To compare, [29], in a large meta-study of educational effects, found that increasing student performance by one standard deviation (SD) typically takes two to three years of schooling, and that increasing an average performing students achievement by one SD means outperforming around 80% of students. Two standard deviations, therefore, means many years of training and outperforming almost all in a regular setting, similar to the numbers reported by [28]. It is expected that such positive effects will be applicable on students achievements in individualized AI-based tutoring, similarly to teacher-based tutoring.

*D. The state of the art (SotA)*

Technology tools for teaching Maths have evolved over the years, from computer algebra systems of the 1980s to AI systems of today [30]. The symbolic manipulators made it possible for symbolic calculations to be done by computers in a much faster way than humans. NLP, especially LLMs, can assist in providing personalized, interactive, and adaptive learning experiences that align closely with the core principles of active learning [31]. In recent years, LLMs have gained significant prominence in different domains and fields. These foundation models are being adapted as accessible chat assistants, such as the SotA DeepSeek [32], OpenAI's ChatGPT [33], Anthropic's Claude [34], and Google's Gemini [35]. Many of the models are based on the deep architecture of the SotA Transformer [36]. Attempts have been made in developing AI tools based on these recent LLMs for the purpose of teaching Maths. Some existing AI tools that are based on these SotA LLMs use traditional non-interactive explanation methods and may be non-user-friendly, thereby making explanations more complicated, e.g. WhyBot[4] and MathGPTPro.[5] MEGA was designed to address these shortcomings.

With regards to Maths explanations, CoT is the SotA and the most intuitive way of giving explanations. There are, however, other methods such as tree-of-thought (ToT), graph-of-thought (GoT), and layer-of-thought (LoT). *Thought* in these methods refer to a single step or concept in the whole process. ToT is based on prompt branching at any given point, to explore different options. The method has the *thought generator* and the *state evaluator*. The thought generator constructs a given number of new nodes, $n$. The state evaluator then generates scores for each. The basis of extending the tree is dictated by the search algorithm in use (e.g., breadth-first-search (BFS) or depth-first-search (DFS)). GoT is based on arbitrary reasoning dependencies among generated thoughts. Each thought can generate multiple child thoughts, as well as have multiple parents. Fianlly, LoT is based on constraint hierarchies to filter and refine candidate responses to a given query to an LLM. The different thought methods are best suited for different problems. For example, CoT, as mentioned earlier is most suitable for Maths and problems with clear sequential reasoning paths, while ToT and GoT are more suitable for complex problems involving multiple reasoning paths. In this work, we restricted the comparison of our experiments and evaluations of MEGA to CoT because of the complexities and cost limitations of human evaluation of additional methods and since CoT is the SotA for Maths.

*E. Ethics of LLM usage in Education*

Users, including educators, students, and guardians, should be aware of the strengths and limitations of LLMs. Providing clear information about how the model functions, its training data, and the criteria for evaluating its outputs promotes trust and accountability. Although LLMs have the potential to enhance learning by offering personalized educational experiences, there is a risk in some cases that excessive reliance on it could weaken students' critical thinking abilities. This risk, in addition to others (e.g. bias, fairness, and excessive energy consumption [37]), raise broader ethical concerns about the role of technology in learning. It is important to carefully assess the ethical consequences of such tools in education and to ensure that they support, rather than replace, the essential role of human educators. Using LLMs as supplementary tools can encourage independent learning and foster critical thinking skills. This is something MEGA aims for with its simplified gamification for students' active learning.

III. METHODOLOGY

*A. LLM System Instruction*

The MEGA method is based on the system instruction (or prompt) provided to a capable LLM and designed to be detailed and descriptive. This final system prompt was arrived at after prompt engineering that involved 2 iterations. Different parts of the instruction (below) capture different important features and are represented by color codes for system prompt indicator (the special characters and term to indicate the type of instruction to the LLM), persona (the character the LLM should assume), chain-of-thought (CoT) (the step-by-step process), Socratic method (the sub-questions), simplified gamification (the pair of answer choices), and formative feedback. Hence, MEGA comprises of the Socratic method, in addition to (1) chain-of-thought reasoning, (2) simplified gamification of 2 answer choices and (3) the formative feedback. In traditional Socratic method, only sub-questions are involved. However, with MEGA, the latter 3 elements are included. We leave the number of sub-questions per main question to be determined by the LLMs and so do not restrict them. Figure 1 is a general representation of the method from which our specific case uses 2 answer choices. MEGA's formative feedback is dependent on the user's choice, as this determines the details of the feedback.

---

[4] whybot-khaki.vercel.app
[5] mathgptpro.com

###Instruction: Following is a request. If the request is a math question, be a helpful assistant and guide the user in a step-by-step detailed explanation before revealing the answer at the end of the steps.
Step 1: At each step of your explanation, ask the user a question with 2 answer choices then wait for the user to make the right choice before going on to the next step of the explanation. Do not go to the next step until the user has answered with a choice.
Step 2: If the user makes the wrong choice then provide the right answer choice before continuing to the next step of the explanation.
Step 3: At the end of it all, ask if the user understands it and if the answer is no then repeat everything from the beginning with an easier explanation.

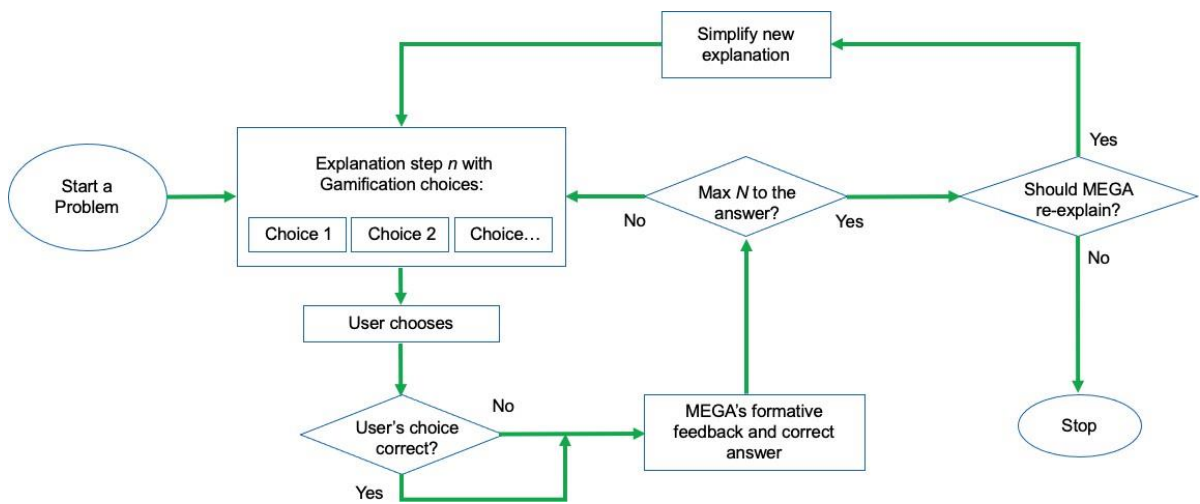

Fig. 1: MEGA method flowchart. Our specific case uses 2 simplified gamification choices.

B. Evaluation Data

We use two different benchmark datasets for our evaluations: Grade School Math 8K (GSM8K) [13] from OpenAI and Mathematics Aptitude Test of Heuristics (MATH) [14]. GSM8K contains Maths word problems and 1,000 samples in the test set. MATH contains problems of Algebra, Number Theory, Probability, and Geometry, among others and 5,000 samples in the test set. The 2 datasets were carefully selected to assess difficulty levels for cases of easy (GSM8K) and relatively hard (MATH) questions for Master students at the University level of education (who are not majoring in Maths). Indeed, [14] had shown that the worst and best performances are 40% and 90%, achieved by a computer science PhD student and International Mathematical Olympiad (IMO) gold medalist, respectively, indicating it's relatively challenging. In their work, it involved 20 randomly selected questions from the MATH test set. More specifically for the MATH dataset, there are 5 difficulty levels from '1' (easiest) to '5' (hardest). We randomly selected 60 questions as the sample size from the test sets of each of the datasets, resulting in a total of 120 samples for evaluation in zero-shot inference (without training). Our decision of 60 from each is based on having (1) a manageable number for student evaluators, (2) a confidence level of 90%, and (3) an error margin of 11%. Examples from each dataset are provided in Table I. There are 2, 11, 19, 12, and 16 problems for levels 1, 2, 3, 4, and 5, respectively, in the samples from the MATH dataset.

C. Models

We evaluated 6 SotA models for capability to follow our carefully-designed instruction and only Generative Pretrained Transformer 4o (GPT4o), Claude 3.5 Sonnet, and Large Language Model Meta AI-3.1-70B-Instruct-Turbo were successful.

TABLE I: Samples per dataset of GSM8K and MATH

| | GSM8K data | | | |
|---|---|---|---|---|
| | **question** | | | **answer** |
| 1 | John rents his car out 10 times a month for 3 hours each time. He gets paid $25 an hour. If his car payment is $500, how much profit does he make on his car? | | | He rents his car 10*3=<<10*3=30>>30 hour a month So he makes 25*30=$<<25*30=750>>750 a month That means he has a profit of 750-500=$<<750-500=250>>250 a month #### 250 |
| 2 | I am three years younger than my brother, and I am 2 years older than my sister. My mom's age is one less than three times my brother's age. When you add all our ages, you get 87. How old am I? | | | Let X be my age. My brother is X + 3 years old. My sister is X - 2 years old. My mom is (X + 3) * 3 - 1 years old. The sum of our ages is X + (X+3) + (X-2) + [(X+3)*3 - 1] = 87 years old. Multiplying through the parentheses and combining like terms, you get X*6 + 9 = 87 years old. Subtracting nine from both sides and dividing both sides by 6, you get X = 13 years old. #### 13 |
| | MATH data | | | |
| | **problem** | **level** | **type** | **solution** |
| 3 | Markov plays a game for three turns. On each turn, he either rolls a fair, six sided die or flips a fair coin. If he rolls a 1 or 2 on the die, he will switch to the coin on the next turn, and if he flips a tails on the coin, he will switch to the die on the next turn. If Markov starts by rolling the die, what is the probability that he will flip the coin on the third turn? | Level 4 | Counting & Probability | We can solve this problem by dividing it into cases. If Markov rolls a 1 or 2 on the first turn, he will flip a coin on the second turn. He must flip a heads to flip a coin on his third turn. There is a $\frac{2}{6} \cdot \frac{1}{2} = \frac{1}{6}$ chance of this case happening. If Markov does not roll a 1 or 2 on the first turn, he will roll the die on the second turn. He must roll a 1 or 2 on the second turn to flip a coin on the third turn. There is a $\frac{4}{6} \cdot \frac{2}{6} = \frac{2}{9}$ chance of this case happening. The total probability that Markov will flip a coin on the third turn is then $\frac{1}{6} + \frac{2}{9} = \boxed{\frac{7}{18}}$. |
| 4 | Carson flips over the cards of a standard 52-card deck one at a time. What is the probability that he flips over the ace of spades before any face card (jack, queen or king)? | Level 5 | Counting & Probability | There are 12 face cards, three from each suit. Within the deck, the 13 relevant cards (the face cards and the ace of spades) are arranged in some order. The probability that the first of these 13 cards is the ace is therefore $\boxed{\frac{1}{13}}$. |

Others were Mistral Large, Cohere command-r-plus-08-2024 (Cohere), and Gemini 1.5 Flash (Gemini). It is worth noting that some of these models have multimodal capabilities (i.e. GPT4o, Claude & Gemini). We used the default hyperparameters of *temperature* = 1 and *top p (or nucleus sampling)* = 1 (or 7 for Claude 3.5 Sonnet). Temperature determines the diversity of the generated responses and top p determines the set of tokens to sample from. Hence, a temperature of 1 allows the maximum diversity in the generated responses. A top p of threshold 1 or 0.7 selects tokens whose cumulative probability equals 1 (all tokens) or 0.7 (most tokens) from which it samples. Equations 1, 2, and 3 are the accuracy metrics we report for each model performance.

$$\text{MEGA Accuracy} = \frac{\text{total correct predictions by MEGA}}{\text{total number of predictions by MEGA}} \quad (1)$$

$$\text{Socratic Accuracy} = \frac{\text{total correct subanswer predictions by MEGA}}{\text{total number of subanswer predictions by MEGA}} \quad (2)$$

$$\text{CoT Accuracy} = \frac{\text{total correct predictions by CoT}}{\text{total number of predictions by CoT}} \quad (3)$$

*D. Human Evaluation*

We held an initial workshop with ML experts to gather input on how best to conduct the evaluation and what additional features to capture (e.g. gender and bias). Besides the model accuracy scores, we perform qualitative human evaluation. Student evaluators (or participants) performed *pairwise comparison* of 2 models, following a blind comparison approach to minimize bias. We use the pairwise comparison approach similarly to [38], who, among others, observed that pairwise comparison is (1) intuitive, (2) becoming increasingly popular in educational measurement, and (3) leads to more accurate evaluations of quality than in cases of one by one evaluations. Besides, a genuine control is arguably impossible for methods that use control groups because of practical difficulties in separating groups since there are many variables involved in such measurement.

More information about the participants is provided in the following sub-section. More specifically, their evaluation involved the pairwise comparison of the traditional CoT (A) and MEGA (B) approaches for (1) Claude 3.5 Sonnet and (2) GPT4o. Generally, 15 different questions from the samples were assigned to a pair of students and the final annotation was the (unanimous) agreement of the two participants. "None" indicates disagreement. A student was assigned the same 15 samples for both methods (A and B), which was executed by the same LLM, and asked to choose "Which explanation was better to understand?" Since "better" may be considered an ambiguous or subjective term, the participants were given an open field at the end of the Microsoft Forms (the evaluation tool) to input a free-form text describing reasons/examples of why A or B is better (e.g., easier or more engaging to understand). The detailed feedback from the students show strong agreement with their choices and provides students' perspective as added value, despite the possibility of subjectivity. Based on the input from the ML experts, to capture any bias on the mode of learning of the evaluators, we also asked at the end of the Microsoft Forms, what type of learning mode (interactive or non-interactive) they usually preferred. The survey was anonymous but gender information was collected. The instruction to the student evaluators is given below. The final question (18) is about their gender and offers the following 3 choices: male, female or prefer not to say [39].

**Instruction:**
For each of the 15 questions below, read the question and the solution explanation of system A (please, ZOOM in if needed). Then copy each question to system B (link below) and interact with it till the end. After that, choose which explanation was better.
B: https://www.megamath.se (You have the login details)

**Question 16:**
Write a short reason (or examples) why you say A or B is better in some/all cases (e.g. easier or more engaging to understand).

**Question 17:**
Usually, which type of learning do you prefer? (Interactive or Non-interactive)

*Participants:* The participants were Master students of the Text Mining course (course code: D7058E) of Luleå University of Technology, Sweden, for the 2024/25 calendar period. Their enrollment in the course automatically made them participants because the evaluation was designed as one of the tasks of the course. A total of 34 students were assigned the task and awarded credits if the task was successfully completed. On average, it took a student about 62 minutes and 61 minutes to complete the evaluation for the GSM8K and MATH datasets, respectively.

IV. RESULTS AND DISCUSSION

*A. Quantitative Results*

Students' evaluations reveal, as shown in Figure 2a, that MEGA's explanations are considered better to understand for their learning experience for both datasets, though only slightly better than CoT (26.67% compared to 24.17% or an average of 16 compared to 14.5 out of 60 instances) for the GSM8K dataset while it is much better than CoT (47.5% compared to 26.67% or an average of 28.5 compared to 16 out of 60 instances) for the MATH dataset. This implies MEGA is better for understanding hard Maths problems compared to easy problems. In Figure 2b, we can observe that the percentage of students' unanimous agreement is higher for GPT4o than Claude 3.5 Sonnet, implying GPT4o is better at its MEGA explanations. In both Figures, heterogeneous quantities make tall error bars more likely (i.e. averaging over different LLMs in Figure 2a and different datasets in Figure 2b). Furthermore, from Figure 2a, we can also notice that the percentage of cases for "None" (where there's no agreement on which method is better) reduces considerably for the MATH dataset. Despite the higher score for MEGA on the MATH dataset, it does not achieve the possible upper-bound of 77.78% of the preferred learning method (interactive) by the students, as indicated in Figure 3. In particular, Figure 3 shows the percentage of student evaluators (along gender categorisation for the two datasets) who usually prefer interactive learning and those who usually prefer non-interactive learning. We observe a higher overall percentage for interactive learning and different ratios across the categories to those who prefer non-interactive learning.

While we may think students generally prefer the interactive learning approach, we were surprised to find out that there are those who prefer the non-interactive approach and by a sizable percentage (22.22%). Part of the reason for MEGA not achieving the possible upper-bound of the preferred learning method is shown in Figure 4, as the LLMs sometimes hallucinate with errors, including at the Socratic (sub-question) steps. They may be correct at some initial steps, wrong at the latter steps and arrive at an incorrect final answer or be incorrect at some steps yet arrive at a correct final answer. Hence, there are slight differences in the MEGA accuracies and the Socratic (sub-question) accuracies. More information is provided on this in Section

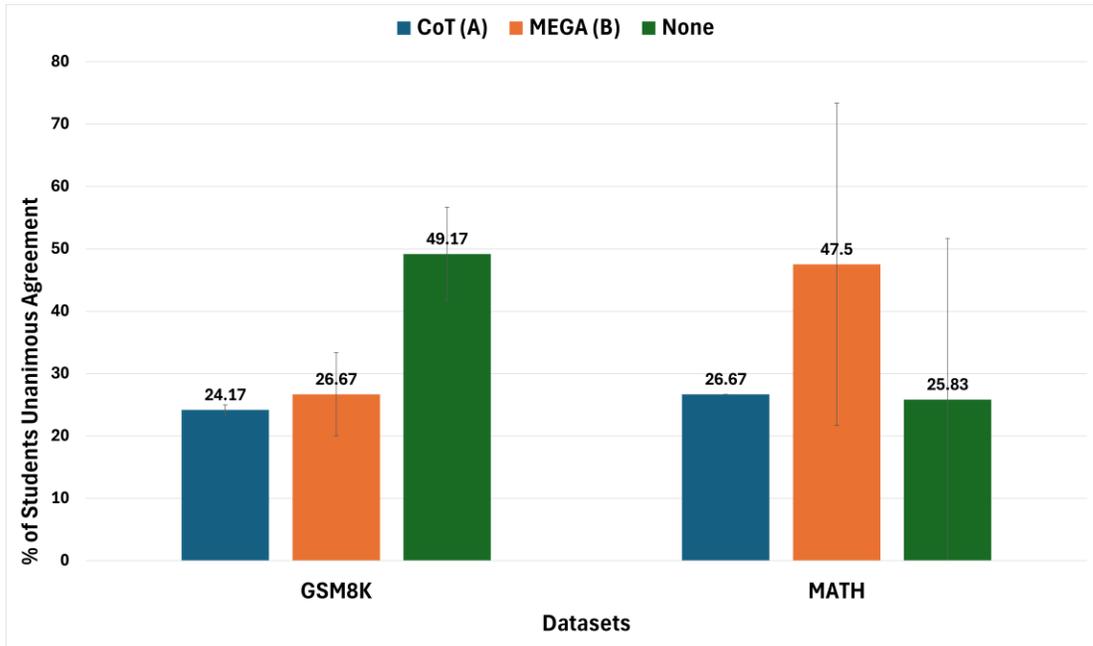

(a) Percentage of unanimous agreements by students for both datasets across the LLMs (Heterogeneous quantities responsible for tall error bars).)

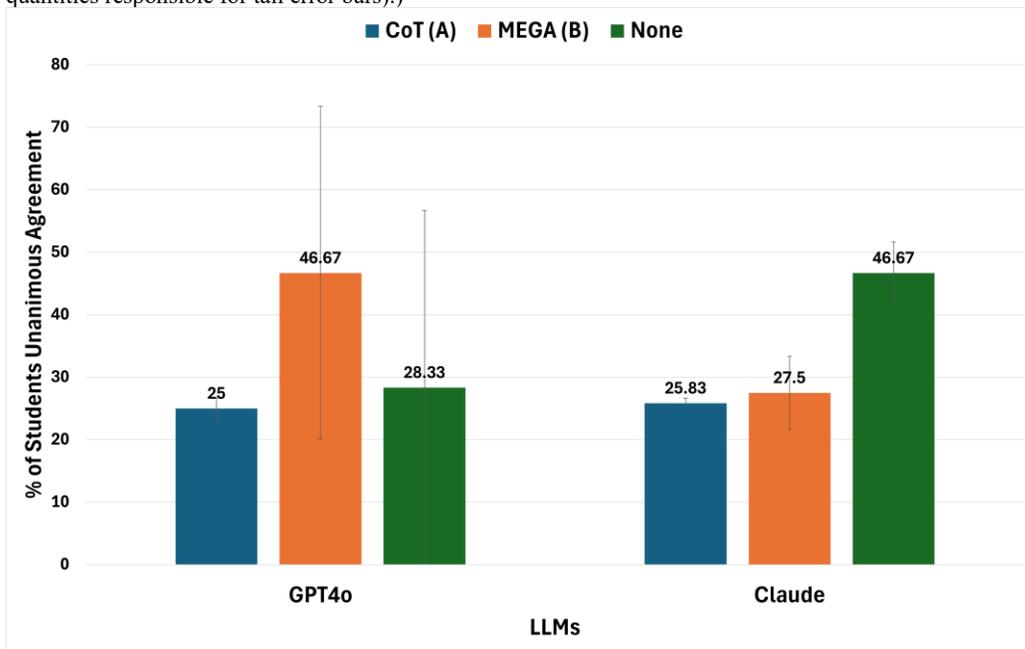

(b) Percentage of unanimous agreements by students for both LLMs across the datasets (Heterogeneous quantities responsible for tall error bars). GPT4o is better.

Fig. 2: Percentage of unanimous agreements

IV-C on analyses. Regarding Figure 3, gender perspectives are important in many studies. They provide additional layer of information and analysis. The figure shows that, for both the male and female genders, there are more students who prefer interactive learning but with different ratios to those who prefer non-interactive learning. It is noteworthy that the minimum, maximum, and average number of sub-questions per main question are 1, 8, and 4, respectively, since we did not fix the number of sub-questions but allowed the LLMs to determine the number. The A/B significance test (at 95% confidence interval and *alpha* of 0.05) shows that, for both cases in Figures 2a and 2b, while the difference is not significant for the GSM8K dataset ($p = 0.6731$), it is significant for the MATH dataset ($p = 0.0201$) and while the difference is not significant for Claude 3.5 Sonnet ($p = 0.4174$), it is significant for GPT4o ($p = 0.0055$), respectively. This implies there may be no real difference in the result for the GSM8K dataset while we are 95% confident that the difference in the percentages (or instances) of agreement are real and not due to chance for the MATH dataset. The same interpretation applies to the LLMs' performance.

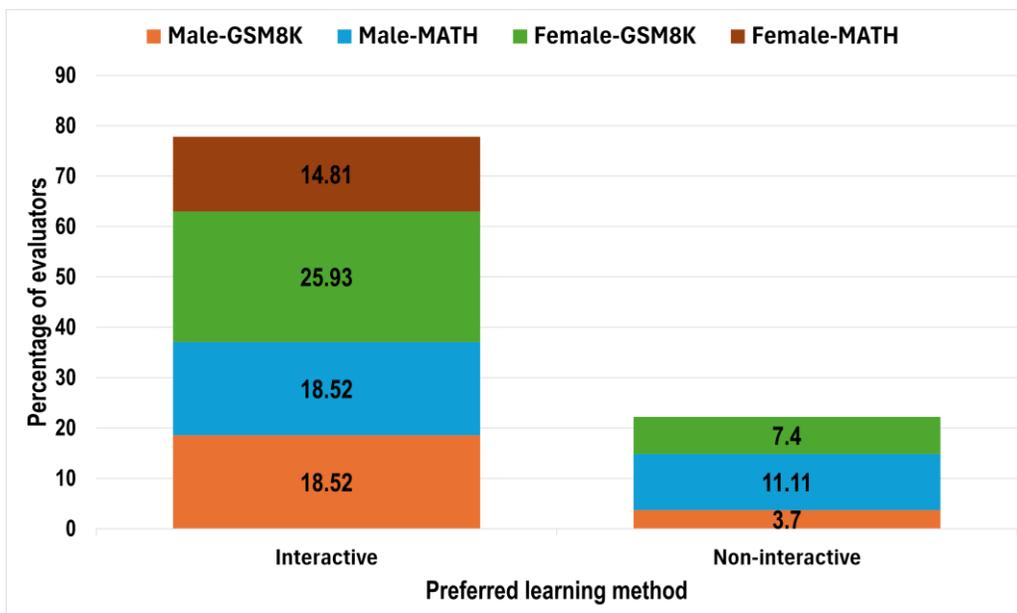

Fig. 3: Preferred learning method of evaluators in percentages, based on gender and dataset.

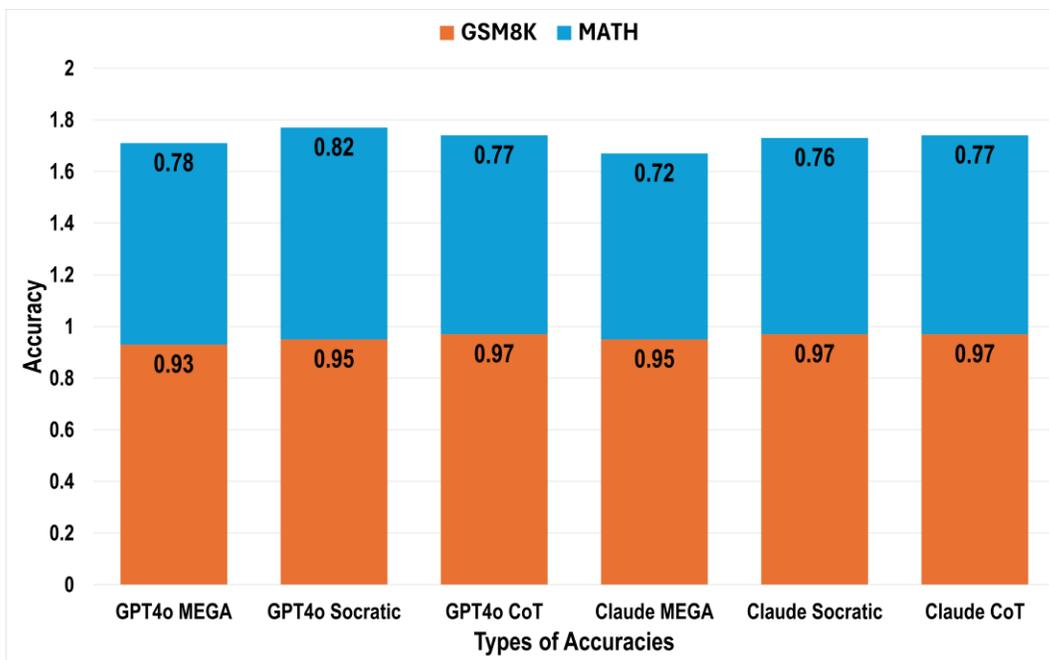

Fig. 4: Average MEGA, Socratic, and CoT accuracies for the two LLMs and the two datasets.

## B. Qualitative Results

One common method of ascertaining if learning has occurred in students is "increased scores in exams" and the MEGA method provides a useful proxy to this through the answer choices provided by the students during the sub-questions. To better understand why MEGA is only slightly better with the GSM8K dataset but much better with the relatively hard MATH problems, we present some examples of the feedback from the students in Table II. Often, the students feel easy questions do not require the additional effort of interactive sub-questions, as this prolongs the learning process unnecessarily - serving as a distraction instead of helping. Meanwhile, the same engaging approach for hard questions provides the level of simplification required to understand an explanation. Based on the survey, the MEGA method is preferred over the CoT for hard questions because (1) it provides more detailed, clearer explanations and (2) it is interactive. We provide qualitative analyses through the use of **content analysis** for all the students' feedback to generate wordclouds of the most frequent key phrases used in the feedback for the two methods, where subfigures 5a, 5b, 5c, and 5d represent CoT-GSM8K, MEGA-GSM8K, CoT-MATH, and MEGA-MATH wordclouds, respectively.

TABLE II: Some evaluation comments (Preferred bias: I- Interactive, N- Non-interactive & Model B- MEGA-based, Model A- CoT-based).

| | Evaluation feedback | I/N | Gender |
|---|---|---|---|
| | GSM8K | | |
| 1 | The questions are very simple and do not require an interactive session, as they mostly involve basic arithmetic operations such as addition and subtraction. Secondly, during the interaction, I noticed that most of the time correct answer is the first option (option 'a'), which conditions the user to select the first option as the correct one. I would definitely prefer an interactive session for complex scenarios where it can explain each question step by step. | I | M |
| 2 | ...However, in other cases, option A was more beneficial because it was simpler and allowed me to grasp the solution faster. For questions like 2, 4, and 11, where the steps were straightforward and easy to follow, option A's direct approach was more efficient. It provided a clear breakdown of the solution without unnecessary interaction, saving time and making it easier to understand without additional questioning. The simplicity of option A allowed me to quickly see all the necessary steps, while option B's repeated straightforward questions sometimes felt like a distraction when the problem was already easy to understand. | I | F |
| 3 | For me, A is better for building understanding because it aligns with the way I think. A step-by-step approach,... B, can sometimes lead to confusion or small errors if not carefully structured. In Q14, the system in B led to the wrong result due to an unclear or poorly organized step. | N | F |
| 4 | B is more engaging and is extremely easy to follow through. Even though A is broken down into smaller parts, it's hard to follow through. B lets me go through step by step in an engaging way and I appreciate that it makes sure that you get each step correct before moving on, this helps you know which step you may need help at. | I | F |
| | MATH | | |
| 5 | In problem 1, neither make any sense to me. I do not understand the task. When I run this in model B, it also makes mistakes in the reversing or doesn't explain correctly. Model A is even worse as there are not much details or explanations and it is not interactive. Thus I can't ask much questions. In all tasks model B shows much clearer the steps with explanations whereas model A is not very understandable for non mathematicians and it is full of extra symbols which makes the text hard to read. Model B is interactive and you can ask for clarification which makes it much better and more advanced. | I | F |
| 6 | B looks more interactive given that it gives one choices on what path to follow. However, I feel it is tedious, I seem to lose interest as the steps are too long sometimes. | N | M |
| 7 | B is more interactive which forces you to understand in a different way. Just reading complex equations can feel overwhelming, and B breaks it down in more managable parts and makes sure you understand before you move on. However, all of the questions where hard to understand in my oponion, in both A and B. Sometimes A is more straight-forward when providing explanations for shorter questions. Also, when B ask you questions it can be hard to know the answer so you still dont really understand anyways. With A you can read it alltogether which can be more understandable at first. Bot overall, B is a more fun way to learn. | I | F |
| 8 | When A is better than B, I generally experience it for more simple math questions, such as 1, 9, 10, 12, 13, and 14. When the question is more challenging/complex, an interactive (option B) is generally better because it is broken down into steps and also asks us to give input. To give input, we first must read and understand the explanation it gives. | I | M |

## C. Analyses

As mentioned earlier, the LLMs hallucinate with errors at the Socratic steps in some instances. Although the differences may be small, we can observe from Figure 4 that comparing the Socratic accuracies to MEGA accuracies we obtain better Socratic scores in all the 4 cases (across the two LLMs and the two datasets). This suggests using the Socratic method of sub-questions can help to improve model accuracies. However, this appears to also expose a drawback in the LLMs because when we compare the Socratic accuracies to the CoT accuracies in all the related 4 cases, we find mixed results. This suggests errors or hallucinations at the Socratic level may sometimes impact negatively on its performance compared to the CoT method.

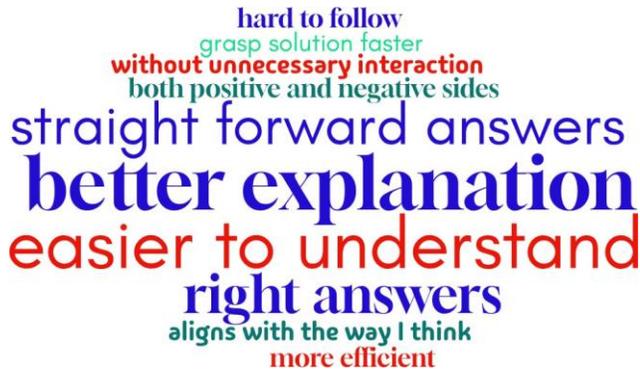

(a) CoT-GSM8K analysis wordcloud with 10 key phrases (*better explanation* is the most frequent phrase, followed by *easier to understand*.)

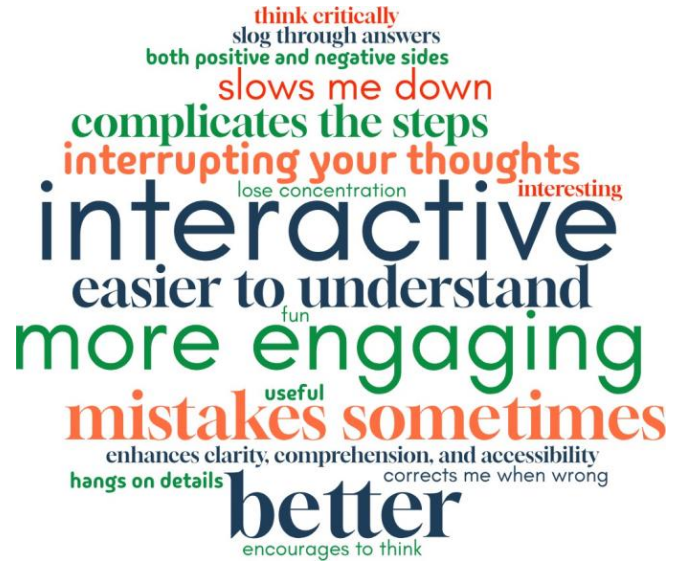

(b) MEGA-GSM8K analysis wordcloud with 19 key phrases (*more engaging* is the most frequent phrase, followed by *interactive*.)

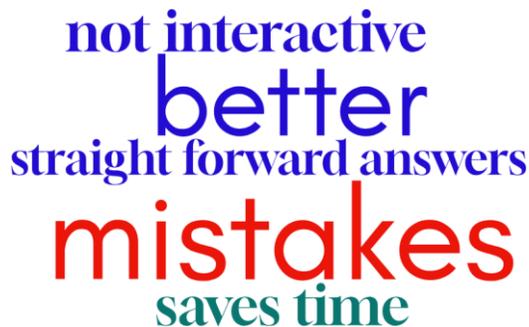

(c) CoT-MATH analysis wordcloud with 5 key phrases (*mistakes* is the most frequent phrase, followed by *better*.)

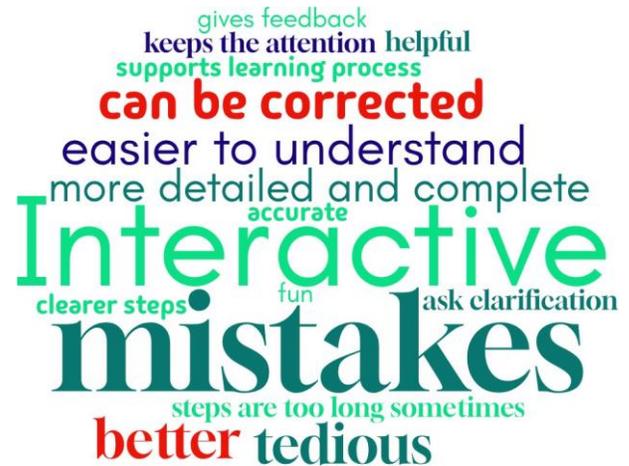

(d) MEGA-MATH analysis wordcloud with 16 key phrases (*interactive* is the most frequent phrase, followed by *mistakes*.)

Fig. 5: Wordclouds of feedback

TABLE III: Average accuracies (and standard deviations) across difficulty levels. Accuracies generally fall as the difficulty rises after level 3.

| Dataset/Level | Frequen y | MEGA Main Acc | Socratic Acc | CoT Acc |
|---|---|---|---|---|
| GSM8K dataset | 60 | 0.94 (0.01) | 0.97 (0.01) | 0.97 (0) |
| MATH dataset | | | | |
| 1 | 2 | **1** (0) | **1** (0.14) | **1** (0) |
| 2 | 11 | 0.86 (0.36) | 0.87 (0.33) | 0.73 (0.45) |
| 3 | 19 | 0.97 (0.16) | 0.99 (0.05) | 0.89 (0.31) |
| 4 | 12 | 0.58 (0.49) | 0.70 (0.39) | 0.58 (0.49) |
| 5 | 16 | 0.50 (0.49) | 0.53 (0.46) | 0.75 (0.43) |

TABLE IV: Average accuracies (and standard deviations) across MATH problem types. Prealgebra questions, followed by Algebra have the highest and next highest accuracies, respectively.

| Type | Frequen y | MEGA Main Acc | Socratic Acc | CoT Acc |
|---|---|---|---|---|
| Intermediate Algebra | 9 | 0.44 (0.50) | 0.54 (0.46) | 0.61 (0.51) |
| Algebra | 29 | 0.83 (0.38) | 0.87 (0.31) | 0.84 (0.36) |
| Counting & Probability | 4 | 0.75 (0.43) | 0.81 (0.35) | 1 (0) |
| Geometry | 6 | 0.75 (0.43) | 0.76 (0.38) | 0.58 (0.49) |
| Prealgebra | 5 | **0.90** (0.30) | **0.90** (0.30) | **0.90** (0.30) |
| Number Theory | 7 | 0.71 (0.45) | 0.74 (0.42) | 0.71 (0.45) |

To have a little insight into what level of difficulty or type of Maths questions are affected, we further analyzed accuracies based on the difficulty levels (presented in Table III) and accuracies based on the type of Maths questions (presented in Table IV). We acknowledge sample numbers at the difficulty level or type in the MATH dataset are merely indicative because of the small number of samples in such cases. We also present example explanations of MEGA and CoT for the same LLM (Claude 3.5 Sonnet) in Chats 1 and 2, respectively, while Chats 3 and 4 are explanations for the same question by MEGA and CoT for GPT4o. Meanwhile, Chats 5 and 6 are additional examples by MEGA-Claude 3.5 Sonnet and CoT-Claude 3.5 Sonnet on geometry from the MATH dataset with level 5 difficulty, where both answers are incorrect (ground truth answer is 16). We can observe from Table III that accuracies generally fall as the difficulty level rises after level 3. Similarly, predictions become less consistent as standard deviations generally rise after level 3. Prealgebra questions appear to have the highest accuracies, followed by Algebra (Table IV). Chat 7 presents an example of a hallucination case for MEGA Claude 3.5 Sonnet on a sample of the GSM8K data while Chat 8 is the CoT method by the same Claude 3.5 Sonnet. In the hallucination example, the LLM makes an error at a latter step of the Socratic process and is corrected by the student before it arrives at the final correct answer. We note that sometimes hallucination is unavoidable as there's no 100% guarantee against it in any existing LLMs. Indeed, human teachers are also prone to errors. In MEGA, however, the inclusion of CoT reasoning and the use of detailed easy explanations make it easier for students to identify, just as in the example provided in Chat 7. Additionally, because many SotA LLMs are trained to be polite they are susceptible to "*sycophancy*" attacks, where the model prioritizes agreement or politeness over accuracy or objectivity. This is another problem which is not 100% solved and can lead to misinformation and bias reinforcement.

## V. CONCLUSION

This work demonstrated that some SotA LLMs are capable of the alternative Maths explanation method of MEGA. They offer better explanations through this method for hard Maths problems compared to the CoT. Hence, MEGA offers students of Mathematics the opportunity to learn using a combination of the Socratic method, CoT, simplified gamification and formative feedback. It leverages the capability of SotA LLMs to deliver personalized student-centered learning. As GenAI advances, we expect more LLMs to be capable of implementing MEGA and to improve their performance, since it is not all SotA LLMs that are currently able to implement the instructions of this method. In future work, the gamification in MEGA may be expanded such that it awards points (or scores) for successfully answered questions. The score may be seen as the measure of understanding a student has for the Maths questions and can be tallied on a leaderboard of competing students. The risk with scoring might be that students may become motivated to earn points instead of actually learning. Furthermore, it will be beneficial to evaluate the learning students acquire through MEGA compared to other traditional explanation approaches. The MEGA method may also be applied to other subjects, besides Maths, to validate the robustness of LLMs in other fields, especially as some other work did not find the Socratic method successful in Physics with ChatGPT. Also, it might be possible to slightly improve the accuracies of the LLMs by setting their temperature to zero, to make the answers more deterministic.

Key take-aways from this study support existing notions or beliefs while some are surprising. The take-aways include: (1) Not all SotA LLMs are capable of the new MEGA method; (2) While most students prefer interactive learning, a sizable percentage (22.22%) usually prefers non-interactive learning, which was a surprise to the authors because some of the participants and students at the initial workshop had expressed the view that **all** students preferred interactive learning, hence, conducting this study helped to debunk that belief; (3) MEGA (and possibly interactive learning, generally) is less preferred when the math problems are simple compared to hard problems, which, again was a bit of a surprise, as the general belief is that there will be equal preference regardless of the difficulty level; (4) Hallucinations can occur at any level of the sub-solutions provided by LLMs, even when the final solution may be correct, or vice-versa. This may be one of the first studies to show this problem.

*Future directions and guidance for the field*

Predicting the future directions of GenAI or LLMs in Maths and signal processing education is not an easy task. The pace of AI development in the last five years has been at an unprecedented speed. The coming years are likely to see developments towards artificial general intelligence (AGI) and artificial super intelligence (ASI), both of which are disruptive technologies. AGI is when a machine or model can perform several multiple (or all) intelligent tasks an average human can perform while ASI is when a machine goes beyond human intelligence in performance [40]. One major weakness of LLMs is the problem of hallucination, however, the use of retrieval augmented generation (RAG) helps to alleviate this problem by providing "factual" context or basis for GenAI to augment its generated output with retrieved information. Similarly, cache-augmented generation (CAG), based on key-value cache without the need for retrieval, can alleviate hallucination. Prompt engineering is another mechanism that is sure to improve in the future. It is a human-friendly mechanism or art of communicating to GenAI in simple natural language with attention on the type of output one desires. Future success in any of the concepts mentioned will certainly lead to success in personalized Maths explanations with MEGA. Also, recommendations by [3] is that the pedagogical approach for teaching Maths-related courses, like signal processing, will benefit by following an interactive and hands-on approach instead of the traditional method.

ACKNOWLEDGEMENTS

The authors wish to thank the Prefekt, Niklas Lehto, Pär-Erik Martinsson, Sabine Mayer, and the Department of Computer Science, Electrical & Space Engineering at Luleå University of Technology for the funding of this project under the SRT.AI initiative. This work was partially supported by the Wallenberg AI, Autonomous Systems and Software Program (WASP), funded by the Knut and Alice Wallenberg Foundation and counterpart funding from Luleå University of Technology (LTU). We also thank the guest editors and the anonymous reviewers for their valuable feedback in enhancing this paper.

**Chat 1: MEGA-Claude's explanation to a MATH problem (3) in Table I**

\>\> User: Markov plays a game for three turns. On each turn, he either rolls a fair, six sided die or flips a fair coin. If he rolls a 1 or 2 on the die, he will switch to the coin on the next turn, and if he flips a tails on the coin, he will switch to the die on the next turn. If Markov starts by rolling the die, what is the probability that he will flip the coin on the third turn?

\>\> MEGA: Let's approach this problem step-by-step:
Step 1: Markov starts by rolling the die. What's the probability that he'll switch to the coin on the second turn?
a) 1/3 (probability of rolling a 1 or 2)
b) 2/3 (probability of rolling a 3, 4, 5, or 6)

\>\> User: a

\>\> MEGA: Correct! The probability of switching to the coin on the second turn is 1/3 (probability of rolling a 1 or 2).
Step 2: Now, let's consider the two possible scenarios for the second turn:
Scenario 1: He switches to the coin (probability 1/3) Scenario 2: He stays with the die (probability 2/3)
For Scenario 1, what's the probability he'll stay with the coin on the third turn?
a) 1/2 (probability of flipping heads)
b) 1/3 (same as the probability of switching from die to coin)

\>\> User: a

\>\> MEGA: Excellent! You're right. If he's flipping the coin on the second turn, the probability of staying with the coin (by flipping heads) is 1/2.
Step 3: For Scenario 2 (staying with the die on the second turn), what's the probability he'll switch to the coin on the third turn?
a) 1/3 (probability of rolling a 1 or 2)
b) 2/3 (probability of rolling a 3, 4, 5, or 6)

\>\> User: a

\>\> MEGA: Correct! The probability of switching to the coin on the third turn, if he's rolling the die on the second turn, is 1/3 (probability of rolling a 1 or 2).

Step 4: Now, let's combine these probabilities. We can use the law of total probability. Which of these correctly represents the probability of flipping the coin on the third turn?
   a) $(1/3 \times 1/2) + (2/3 \times 1/3)$
   b) $(1/3 \times 1/2) + (1/3 \times 1/3)$

>> User: a

>> MEGA: Excellent! You're absolutely correct.
The probability of flipping the coin on the third turn is: $(1/3 \times 1/2) + (2/3 \times 1/3)$
Let's calculate this:
Step 5: What's the result of this calculation?
   a) 5/18
   b) 7/18
>> User: b

>> MEGA: Perfect! Your calculation is correct.
$(1/3 \times 1/2) + (2/3 \times 1/3) = 1/6 + 2/9 = 3/18 + 4/18 = 7/18$
Therefore, the probability that Markov will flip the coin on the third turn is 7/18.
Do you understand the solution, or would you like me to explain any part of it again?

**Chat 2: CoT-Claude's explanation to a MATH problem (3) in Table I**
Let's approach this step-by-step:
1) First, let's consider the probability of switching from die to coin: P(switch from die to coin) = P(rolling a 1 or 2) = 2/6 = 1/3

2) Now, the probability of switching from coin to die: P(switch from coin to die) = P(flipping tails) = 1/2

3) Markov starts by rolling the die. To flip the coin on the third turn, one of two things must happen: a) He stays with the die on the second turn and then switches to the coin. b) He switches to the coin on the second turn and then stays with the coin.

4) Let's calculate the probability of scenario a: P(stay with die on 2nd turn) = 1 - 1/3 = 2/3 P(switch to coin on 3rd turn) = 1/3 P(scenario a) = 2/3 * 1/3 = 2/9

5) Now for scenario b: P(switch to coin on 2nd turn) = 1/3 P(stay with coin on 3rd turn) = 1 - 1/2 = 1/2 P(scenario b) = 1/3 * 1/2 = 1/6 6) The total probability is the sum of these two scenarios: P(flip coin on 3rd turn) = P(scenario a) + P(scenario b) = 2/9 + 1/6 = 4/18 + 3/18 = 7/18

Therefore, the probability that Markov will flip the coin on the third turn is 7/18.

**Chat 3: MEGA-GPT4o's explanation to a MATH problem (3) in Table I**
>> User: Markov plays a game for three turns. On each turn, he either rolls a fair, six sided die or flips a fair coin. If he rolls a 1 or 2 on the die, he will switch to the coin on the next turn, and if he flips a tails on the coin, he will switch to the die on the next turn. If Markov starts by rolling the die, what is the probability that he will flip the coin on the third turn?

>> MEGA: To solve this problem, let's break it down step-by-step.
**Step 1**: Markov starts by rolling a die. The outcomes of the first roll can determine the type of action he takes on the second turn.
If Markov rolls a 1 or 2 on the die, he will switch to flipping a coin on the second turn. If he rolls a 3, 4, 5, or 6, he continues rolling the die on the second turn.
What is the probability that, after the first roll (D1), he switches to flipping a coin (C2)?
   - A) $\frac{1}{3}$ - B) $\frac{1}{2}$
Please choose the correct probability.

>> User: a

>> MEGA: Correct! The probability that Markov rolls a 1 or 2 and switches to flipping a coin on the second turn is $\frac{2}{6} = \frac{1}{3}$.
**Step 2**: Let's now consider the second turn. Depending on what he did in the first turn, there are two scenarios:
   - If Markov is flipping a coin on the second turn (C2): - Flipping heads (H2) means he continues flipping a coin on the third turn (C3). - Flipping tails (T2) means he switches back to rolling the die on the third turn (D3).
   - If Markov is rolling the die on the second turn (D2): - Rolling a 1 or 2 leads to flipping a coin on the third turn (C3). - Rolling a 3, 4, 5, or 6 means he continues rolling the die on the third turn (D3).
What is the probability of continuing to flip a coin if Markov was flipping a coin on the second turn (C2 and then H2)?
   - A) $\frac{1}{2}$ - B) $\frac{1}{4}$
Please choose the correct answer.

>> User: a

>> MEGA: Correct! The probability of flipping heads and continuing to flip a coin on the third turn is $\frac{1}{2}$.
**Step 3**: Let's consider the "die on the second turn" (D2) scenario. If Markov was rolling the die on the second turn:
   - Rolling a 1 or 2 means he switches to flipping a coin on the third turn (C3). - Rolling a 3, 4, 5, or 6 means he continues rolling the die on the third turn (D3).
What is the probability of flipping a coin on the third turn if Markov was rolling the die on the second turn (D2 and then 1 or 2)?

- A) $\frac{1}{3}$ - B) $\frac{1}{6}$

Please select the correct answer.

>> User: b

>> MEGA: Actually, the correct probability is $\frac{2}{6} = \frac{1}{3}$ because rolling a 1 or 2 from a 6-sided die means flipping the coin in the next round.
**Step 4**: We now need to calculate the total probability that Markov is flipping a coin on the third turn ($C3$). We can consider the two paths leading to $C3$:
1. **Path from D1 to C2 to C3**: - Probability of $D1$ to $C2$ is $\frac{1}{3}$. - Probability of $C2$ to $C3$ (given heads) is $\frac{1}{2}$. - Combined probability: $\frac{1}{3} \times \frac{1}{2} = \frac{1}{6}$.
2. **Path from D1 to D2 to C3**: - Probability of $D1$ to $D2$ is $\frac{2}{3}$. - Probability of $D2$ to $C3$ is $\frac{1}{3}$. - Combined probability: $\frac{2}{3} \times \frac{1}{3} = \frac{2}{9}$.
**Step 5**: Add the probabilities from both paths to get the total probability for $C3$:

$$\frac{1}{6} + \frac{2}{9}$$

To add these, find a common denominator. What is the common denominator?
- A) 36 - B) 18
Please choose the correct option.

>> User: b

>> MEGA: Correct! The common denominator is 18.
**Step 6**: Let's convert both fractions to have a denominator of 18: - Convert $\frac{1}{6}$ to have a denominator of 18: $\frac{1}{6} = \frac{3}{18}$. - Convert $\frac{2}{9}$ to have a denominator of 18: $\frac{2}{9} = \frac{4}{18}$
Now, add them together:

$$\frac{3}{18} + \frac{4}{18} = \frac{7}{18}$$

Therefore, the probability that Markov will be flipping a coin on the third turn is $\frac{7}{18}$.
Do you understand how we arrived at this answer?

>> User: yes

**Chat 4: CoT-GPT4o's explanation to a MATH problem (3) in Table I**
Given that Markov starts by rolling a die, we need to find the probability that he will be flipping a coin on the third turn. Let's analyze each turn and the possible outcomes. 1. On the first turn, since he starts with a die (D1), there are two possible outcomes: - Rolls a 1 or 2: transitions to flip a coin (C2) on the second turn. - Rolls a 3, 4, 5, or 6: stays with rolling the die (D2) on the second turn.
The probability of D1 transitions:

$$P(C2) = \frac{2}{6} = \frac{1}{3}$$
$$P(D2) = \frac{4}{6} = \frac{2}{3}$$

2. On the second turn, there are two scenarios – either C2 or D2: - If he is at C2, the outcomes of a coin flip are: - Heads (H2): continues to flip a coin (C3) on the third turn. - Tails (T2): switches to roll die (D3) on the third turn. - If he is at D2, the outcomes of a die roll are: - Rolls a 1 or 2: switches to flip coin (C3) on the third turn. - Rolls a 3, 4, 5, or 6: continues with die roll (D3).
Let's compute the probabilities for both forks: For $P(C2) = \frac{1}{3}$:

$$P(C3 \mid C2, \text{heads}) = \frac{1}{2}$$
$$P(D3 \mid C2, \text{tails}) = \frac{1}{2}$$

For $P(D2) = \frac{2}{3}$:

$$P(C3 \mid D2) = \frac{2}{6} = \frac{1}{3}$$
$$P(D3 \mid D2) = \frac{4}{6} = \frac{2}{3}$$

3. We now calculate the total probability of starting at $D1$ and ending up flipping a coin on $C3$ on the third turn. **Path 1: D1 → C2 → C3**

$$P(D1 \to C2) = \frac{1}{3}$$
$$P(C2 \to C3 \mid \text{heads}) = \frac{1}{2}$$
$$P(D1 \to C2 \to C3) = \frac{1}{3} \times \frac{1}{2} = \frac{1}{6}$$

**Path 2: D1 → D2 → C3**

$$P(D1 \to D2) = \frac{2}{3}$$

$$P(D2 \to C3) = \frac{1}{3}$$

$$P(D1 \to D2 \to C3) = \frac{2}{3} \times \frac{1}{3} = \frac{2}{9}$$

Sum probability paths resulting in C3:

$$P(D1 \to C3) = \frac{1}{6} + \frac{2}{9}$$

Convert fractions common denominator:

$$P(D1 \to C3) = \frac{3}{18} + \frac{4}{18} = \frac{7}{18}$$

Therefore, the probability that Markov will be flipping a coin on the third turn is:

$$\frac{7}{18}$$

**Chat 5: Example by MEGA-Claude 3.5 Sonnet to a geometry MATH problem of level 5**

>> User: The parallelogram bounded by the lines $y = ax + c$, $y = ax + d$, $y = bx + c$, and $y = bx + d$ has area 18. The parallelogram bounded by the lines $y = ax + c$, $y = ax - d$, $y = bx + c$, and $y = bx - d$ has area 72. Given that $a$, $b$, $c$, and $d$ are positive integers, what is the smallest possible value of $a + b + c + d$?

>> MEGA: Let's approach this step-by-step: Step 1: Recall the formula for the area of a parallelogram formed by two pairs of parallel lines $y = mx + k1$ and $y = mx + k2$, and $y = nx + k1$ and $y = nx + k2$. The area is given by: $|k2 - k1| * |n - m|$. Which of these correctly applies this to our first parallelogram?
A) $18 = |d - c| * |b - a|$
B) $18 = (d - c) * (b - a)$

>> User: A

>> MEGA: Excellent! You're correct. For the first parallelogram, we have: $18 = |d - c| * |b - a|$ Now, let's consider the second parallelogram. Step 2: Apply the same formula to the second parallelogram. Which equation represents the area of the second parallelogram?
A) $72 = |d + d| * |b - a|$
B) $72 = |-d - d| * |b - a|$

>> User: B

>> MEGA: Great job! You're correct. For the second parallelogram, we have: $72 = |-d - d| * |b - a|$
Let's simplify this equation. Step 3: Simplify the equation for the second parallelogram.
$72 = |-d - d| * |b - a|$ $72 = |2d| * |b - a|$ $72 = 2d * |b - a|$
Now, let's compare our two equations: $18 = |d - c| * |b - a|$ $72 = 2d * |b - a|$ Step 4: What can we conclude about $|d - c|$ and $2d$?
A) $2d = 4 |d - c|$
B) $2d = 3 |d - c|$

>> User: A

>> MEGA: Excellent reasoning! You're correct. $2d = 4|d - c|$ Now, let's continue our analysis. Step 5: Given that d and c are positive integers, what's the smallest possible value for $|d - c|$?
A) 1
B) 2

>> User: A

>> MEGA: Correct! The smallest possible value for $|d - c|$ is indeed 1. Now, let's use this information to find the values of c and d. Step 6: If $|d - c| = 1$ and $2d = 4 |d - c|$, what are the values of c and d?
A) $c = 1$, $d = 2$
B) $c = 2$, $d = 3$

>> User: A

>> MEGA: Excellent! You're correct. If $|d - c| = 1$ and $2d = 4$, then: $c = 1$ and $d = 2$ Now, let's find the value of $|b - a|$. Step 7: Using the equation $18 = |d - c| * |b - a|$, and knowing that $|d - c| = 1$, what is $|b - a|$?
A) 18
B) 9

>> User: A

>> MEGA: Great job! You're correct. If $18 = 1 * |b - a|$, then $|b - a| = 18$. Now, let's determine the smallest possible values for a and b. Step 8: Given that a and b are positive integers and $|b - a| = 18$, what are the smallest possible values for a and b?
A) $a = 1$, $b = 19$

B) a = 2, b = 20
>> User: A

>> MEGA: Excellent! You're absolutely correct. The smallest possible values are: a = 1 and b = 19 Now we have all the values: a = 1 b = 19 c = 1 d = 2 Step 9: Calculate a + b + c + d. What is the sum of these values?
A) 23
B) 24

>> User: A

>> MEGA: Perfect! You've solved it correctly. a + b + c + d = 1 + 19 + 1 + 2 = 23

Therefore, the smallest possible value of a + b + c + d is 23.
To summarize: 1. We derived that 2d = 4|d - c|2. We found the smallest possible values: c = 1, d = 2 3. We calculated that |b - a|= 18 4. We determined the smallest possible values: a = 1, b = 19 5. We summed a + b + c + d to get 23
Excellent work on solving this complex problem!

**Chat 6: Example of CoT-Claude 3.5 Sonnet to the same geometry problem of Chat 5**
Let's approach this step-by-step: 1) The area of a parallelogram formed by two pairs of parallel lines y = ax + k1, y = ax + k2, y = bx + k1, and y = bx + k2 is given by the formula: Area = |k2 - k1|* |b - a|
2) For the first parallelogram: 18 = |d - c|* |b - a|
3) For the second parallelogram: 72 = |(-d) - c|* |b - a|= (d + c) * |b - a|
4) Dividing the second equation by the first: 72/18 = (d + c)/(d - c) 4 = (d + c)/(d - c)
5) Cross multiply: 4(d - c) = d + c 4d - 4c = d + c 3d = 5c
6) Since a, b, c, and d are positive integers, the smallest possible values for c and d that satisfy this relation are: c = 3 and d = 5
7) Substituting these back into the equation from step 2: 18 = (5 - 3) * |b - a|18 = 2 * |b - a|9 = |b - a|
8) The smallest possible positive integer values for a and b that satisfy this are: a = 1 and b = 10 (or vice versa)
9) Therefore, the smallest possible values are: a = 1, b = 10, c = 3, d = 5 10) The sum a + b + c + d = 1 + 10 + 3 + 5 = 19

Thus, the smallest possible value of a + b + c + d is 19.

**Chat 7: Hallucination example by model B (MEGA-Claude) to a GSM8K problem (Hallucination in bold).**
>> User: Every day, Wendi feeds each of her chickens three cups of mixed chicken feed, containing seeds, mealworms and vegetables to help keep them healthy. She gives the chickens their feed in three separate meals. In the morning, she gives her flock of chickens 15 cups of feed. In the afternoon, she gives her chickens another 25 cups of feed. How many cups of feed does she need to give her chickens in the final meal of the day if the size of Wendi's flock is 20 chickens?

>> MEGA: Certainly! I'll guide you through solving this problem step by step. Let's begin!
Step 1: Let's identify how much feed each chicken needs daily. How many cups of feed does each chicken need per day?
A) 2 cups
B) 3 cups

>> User: B

>> MEGA: Excellent! You're correct. Each chicken needs 3 cups of feed per day. Let's move on to the next step.
Step 2: Now, let's calculate the total amount of feed needed for all of Wendi's chickens daily. How many cups of feed are needed for 20 chickens?
A) 60 cups
B) 40 cups

>> User: A

>> MEGA: Great job! You're right. The total amount of feed needed for 20 chickens is 60 cups (20 chickens × 3 cups each). Let's continue to the next step.
Step 3: We know Wendi has already given her chickens two meals. Let's calculate how much feed has been given so far. How many cups of feed have been given in the morning and afternoon combined?
A) 35 cups
B) 40 cups

>> User: B

>> MEGA: **I apologize, but that's not correct.** Let's review the information:
- In the morning, Wendi gives her chickens 15 cups of feed. - In the afternoon, she gives them 25 cups of feed.
To find the total amount given so far, we add these two amounts: 15 cups + 25 cups = 40 cups
**So, the correct answer is actually A) 35 cups.**
Let's move on to the final step.
Step 4: Now we can calculate how much feed Wendi needs to give in the final meal. Which of these calculations gives us the correct amount for the final meal?
A) 60 cups - 40 cups = 20 cups

B) 60 cups - 35 cups = 25 cups

>> User: That's not right, think again

>> MEGA: I apologize for the confusion. You're absolutely right, and I thank you for catching that error. Let me correct myself:
You are correct that the total amount given so far is indeed 40 cups (15 cups in the morning + 25 cups in the afternoon).
So, let's revise Step 4:
Step 4: Now we can calculate how much feed Wendi needs to give in the final meal. Which of these calculations gives us the correct amount for the final meal?
A) 60 cups - 40 cups = 20 cups
B) 60 cups - 35 cups = 25 cups

>> User: A

>> MEGA: Excellent! You're absolutely correct, and thank you for your patience. Let's confirm the solution:
A) 60 cups - 40 cups = 20 cups is the correct calculation.
Here's a summary of the solution: 1. Each chicken needs 3 cups of feed per day. 2. For 20 chickens, the total daily feed needed is 60 cups (20 × 3). 3. Wendi has already given 40 cups (15 in the morning + 25 in the afternoon). 4. For the final meal, she needs to give: 60 - 40 = 20 cups.
Therefore, Wendi needs to give her chickens 20 cups of feed in the final meal of the day.
Do you understand the solution, or would you like me to explain any part of it again?

**Chat 8: CoT-Claude's explanation to the same GSM8K problem in Chat 7**
Let's break this problem down step by step:
1. Each chicken gets 3 cups of feed per day. 2. Wendi has 20 chickens. 3. The feed is given in three separate meals. 4. In the morning, she gives 15 cups. 5. In the afternoon, she gives 25 cups. 6. We need to find out how much she should give in the final meal.
Now, let's solve:
1. Calculate the total amount of feed needed for all chickens: * 3 cups per chicken × 20 chickens = 60 cups total
2. Calculate how much has been given so far: * Morning meal: 15 cups * Afternoon meal: 25 cups * Total given so far: 15 + 25 = 40 cups
3. Calculate how much is left to give: * Total needed - Total given so far = Amount for final meal * 60 cups - 40 cups = 20 cups
Therefore, Wendi needs to give her chickens 20 cups of feed in the final meal of the day.

## APPENDIX

**MEGA's correct solution of the Laplace transform of the Heaviside function supplied as an image modality input.**

>> User: Solve the problem in the image

>> MEGA: To find the Laplace transform of the unit-step (Heaviside) function $u_s(t)$, we start with its definition:

$$u_s(t) = \begin{cases} 0, & t \leq 0 \\ 1, & t > 0 \end{cases}$$

The Laplace transform of a function $f(t)$ is given by:

$$L\{f(t)\} = \int_0^\infty f(t)e^{-st}\, dt$$

For the unit-step function $u_s(t)$, but can be treated as $f(t) = 1$ for $t > 0$. Let's use this in the equation:

$$L\{u_s(t)\} = \int_0^\infty 1 \cdot e^{-st}\, dt$$

Next, calculate the integral:

$$\int_0^\infty e^{-st}\, dt$$

For this integral, what is the antiderivative of $e^{-st}$?
A) $-\frac{1}{s}e^{-st}$
B) $\frac{1}{s}e^{-st}$

>> User: b

>> MEGA: Correct!

The antiderivative of $e^{-st}$ is $-\frac{1}{s}e^{-st}$.
Now, let's evaluate the integral:

$$L\{u_s(t)\} = \left. -\frac{1}{s}e^{-st} \right|_0^\infty$$

First, evaluate the limit as $t \to \infty$ for the upper bound:

$$\lim_{t \to \infty} \left\{ -\frac{1}{s} e^{-st} \right\} = 0 \quad \text{(since } s > 0\text{)}$$

Now evaluate at the lower bound $t = 0$:

$$-\left\{ -\frac{1}{s} e^{-s \times 0} \right\} = -\left\{ -\frac{1}{s} \right\} = \frac{1}{s}$$

Putting it all together, the Laplace transform is:

$$\mathcal{L}\{u(t)\} = 0 - \left\{ -\frac{1}{s} \right\} = \frac{1}{s}$$

So the Laplace transform of the unit-step function is $\frac{1}{s}$.
Do you understand this approach?

>> User: yes thanks

## Author Biographies

**Tosin Adewumi** (tosin.adewumi@ltu.se) received his PhD degree in Machine Learning (with specialization in Natural Language Processing) from Luleå University of Technology, Sweden, in 2022. He is a postdoc in the same university. He is a Wallenberg AI, Autonomous Systems and Software Program (WASP) faculty member and a member of the German Academic Exchange Service AInet Fellows and Alumni Network (DAAD AInet). He is also a member of the Nigeria Computer Society (NCS). His research interests include Large Language Models, Large Multimodal Models, and AI Ethics. He serves as program committee member/reviewer for conference and journal venues like Empirical Methods in Natural Language Processing (EMNLP).

**Foteini Liwicki** (foteini.liwicki@ltu.se) received her PhD from the School of Electrical and Computer Engineering, Greece, in the field of Pattern Recognition in 2015 with the title "Recognition of online handwritten mathematical expressions". Over two decades at the Institute of Language and Speech Processing (ATHENA R.C.), she coordinated research projects on pattern recognition, NLP, and the creation of interactive educational platforms across Europe. From 2015 till June 2019 she worked as a PostDoc fellow in the University of Fribourg in the field of Document Image Analysis and Database generation. From June 2018 till June 2019 she worked as a PostDoc fellow with the Machine Learning group at the Luleå University of Technology, Sweden. Her research explores how AI can transform learning experiences and support industries, such as drill core analysis, while investigating how humans process language and thought.

**Marcus Liwicki** (marcus.liwicki@ltu.se) received his PhD degree from the University of Bern, Switzerland, in 2007, and his habilitation degree at the Technical University of Kaiserslautern, Germany, in 2011. Currently he is chaired professor in Machine Learning and vice-rector for AI at Luleå University of Technology. His research interests include machine learning, pattern recognition, artificial intelligence, human computer interaction, digital humanities, knowledge management, ubiquitous intuitive input devices, document analysis, and graph matching. From October 2009 to March 2010 he visited Kyushu University (Fukuoka, Japan) as a research fellow (visiting professor), supported by the Japanese Society for the Promotion of Science. In 2015, at the young age of 32, he received the ICDAR young investigator award, a bi-annual award acknowledging outstanding achievements of in pattern recognition for researchers up to the age of 40.

**Viktor Gardelli** (viktor.gardelli@ltu.se) is an Associate Professor in Education specializing in Philosophy at the department of Health, Learning and Technology at Luleå University of Technology. His research interests include dialogic education, including philosophical and socratic dialogs, ethics, logic, argumentation, philosophy of education, AI in education, and philosophy education. He's worked on projects funded by Vinnova and Vetenskapsrådet, including using dialogic education to support people with aphasia and persons with acquired brain injury. Currently, he's involved in the AI4EDU project funded by the EU, and as a researcher in ArcTech Learning Lab, a research center for digital learning.

**Lama Alkhaled** (lama.alkhaled@ltu.se) is a project manager specializing in artificial intelligence at ProcessIT Innovations, Luleå University of Technology (LTU) in Sweden. With a master's degree in computer science and a postgraduate certificate in business studies, she combines technical expertise with strategic leadership. Lama manages several national and EU-funded projects that aim to integrate AI into regional industries and SMEs. Lama contributed as a research engineer in LTU's ML Group, focusing on AI solutions for mental health. Lama's research spans machine learning, natural language processing, and image processing, with publications in international journals and conferences. She has strong leadership, multitasking, and communication skills.

**Hamma Mokayed** (hamam.mokayed@ltu.se) is an Associate Professor at Luleå University of Technology, specializing in machine learning within the Embedded Intelligent Systems Lab. With over 15 years of experience in the vehicle intelligence industry and academia. His expertise encompasses computer vision, machine learning, and deep learning, particularly in vehicle intelligence systems and AI applications in education. He has also completed more than 20 credits in higher education didactics and pedagogy. Also he is a team member in the AI4Edu project.